\documentclass[11pt]{article}
\usepackage[T1]{fontenc}
\usepackage{graphicx}
\usepackage{booktabs}
\usepackage[utf8]{inputenc}
\usepackage{amsmath}
\usepackage{amssymb}
\usepackage{latexsym}
\usepackage[english]{babel}
\usepackage{hyperref}
\usepackage{xcolor}
\usepackage[left=2.5cm,right=2.5cm]{geometry}
\usepackage[misc]{ifsym}  
\usepackage{hyperref}
\usepackage{orcidlink}

\begin{document}

\title{Possibility-based Inference: A Deterministic Framework for Robust and Reliable Uncertainty Quantification in Decision-Making}

\author{Seyedfakhredin Musavishavazi\,\href{mailto:sfmusavi@gmail.com}{\Letter}\,\orcidlink{0000-0001-6494-6258} \and
Maryam Alipour}

\date{}

\maketitle

\begin{abstract}
Decision-making under uncertainty remains a key issue in Machine Learning (ML), particularly in the deployment of safety-critical systems, where reliable quantification of system behavior is essential for evaluating behavioral impacts and strengthening confidence in ML models. This contribution, influenced by fuzzy set theory and grounded in Bayesian statistics, addresses challenges related to data precision and the performance–accuracy trade-off in Bayesian inference (BI). To this end, we introduce a possibility-based contrastive grade of prediction as an alternative quantification method to existing latent posterior-based approaches for parameter estimation (approximation) and model selection. Unlike conventional Bayesian inference, which represents uncertainty within the probabilistic framework, the proposed model adopts a possibility-based representation that allows uncertain and potentially conflicting information to be characterized more flexibly. Rather than focusing on isolated aspects of a problem, the resulting mathematical framework provides a systematic approach to data mining while mitigating overly confident, yet implausible predictions generated by highly specialized deep learning architectures.

\end{abstract}

\section{Introduction}
In the context of ML, uncertainty refers to the lack of confidence in the results produced by a model. 
Looking at the notion of generalization beyond the data seen during the learning step, it is important to recognize that any decision made by an ML model is not provably certain but rather hypothetical \cite{E63B}. 
Consequently, accounting for uncertainty is essential in ML models to ensure \textbf{robust} inference \cite{M20R,M19R}, facilitating the transition from real-world data to reliable predicted results.
Several factors can influence the decisions made by a model and create uncertainty, such as insufficient or biased data, noise in the data \cite{Z04C}, model architecture, and lack of \textbf{interpretability} \cite{A21A,M20A}, to name a few. 
Taking these factors into account and mitigating them improves uncertainty and increases the \textbf{trustworthiness} of model decisions \cite{H23M}. 
\newline
\newline
\noindent
In statistical terms, uncertainty is encapsulated through standard probabilities and probabilistic predictions. The model’s ability to generalize beyond the data it has learned from is a fundamental requirement, achieved through the process of induction, where the model infers patterns and rules based on observed data and applies them to unseen data. To this end, it is assumed that the training and test data are identically and independently distributed (i.i.d.). However, this assumption often does not hold in practice, leading to a deterioration in model predictions due to domain distribution gaps. In addition, collecting all possible data to train the model is expensive and practically impossible. As a consequence, the predictions made by such a model are inherently uncertain \cite{W22G}. For this reason, the uncertainty in mathematical models stems from the \textbf{stochastic} perspective of the model’s output, in contrast to a \textbf{deterministic} output \cite{N03M}.
\newline
\newline
\noindent
There are two main sources of uncertainty: data and lack of knowledge. Data uncertainty, also known as aleatoric uncertainty, is linked to the stochastic nature of the data generation process, making it impossible to eliminate completely \cite{D09A}. In contrast, knowledge uncertainty, or epistemic uncertainty, arises from information gaps and can potentially be reduced by employing state-of-the-art models or gathering additional data \cite{G20E}. Epistemic uncertainty encompasses both model uncertainty and approximation uncertainty: model uncertainty pertains to the construction of the model, while approximation uncertainty relates to parameter estimation \cite{H23M}.
\newline
\newline
\noindent
Model uncertainty is commonly addressed through the use of over-parameterized neural networks \cite{L18L,R23O,V19F,Y23A,Z21U}, particularly within the framework of Bayesian Neural Networks (BNNs) \cite{A24U,C10A,J24A,P24T,R16B,S20D}. BNNs represent model parameters as random variables governed by a posterior probability distribution, thereby enabling principled uncertainty quantification. However, exact posterior inference is generally intractable, which motivates the use of Variational Inference (VI) \cite{J99A,W08G}, a deterministic approximation approach to Bayesian inference. VI seeks to approximate the true posterior with a tractable distribution by minimizing a divergence measure, most commonly the Kullback–Leibler (KL) divergence \cite{K51I}, between the two distributions.
Through this optimization, VI identifies the optimal member of a predefined variational family that best approximates the posterior distribution induced by the observed data.
\newline
\newline
\noindent
By providing improved accuracy while analytically reducing computational complexity relative to the original model, this mathematical and statistical framework enables more reliable decision-making and is particularly well suited for data assimilation (DA) \cite{C18D,G07A,H98D,H01A,O04A,S06S}, which integrates real-world observations into dynamic models to estimate the true state of complex systems and to generate task-specific neural networks with real-valued outputs.
\newline
\newline
\noindent
However, the development of task-specific models for practical deployment, together with the scalability challenges posed by model architectures, has brought computational demands on the one hand and uncertainty in data precision on the other to the forefront, highlighting the need to balance trade-offs between performance and accuracy \cite{D24A,V18S}. Consequently, this motivates the use of a possibilistic approach, rather than a probabilistic one, as a suitable alternative framework for information analysis \cite{Z99F}.
\newline
\newline
\noindent
To motivate this approach, exploring the concept of possibility, and especially its interconnection with other reasoning frameworks, provides a basis for establishing a robust conceptual platform \cite{M80N,Z65F}. This platform is particularly advantageous for addressing challenges where imprecision arises from the lack of precisely defined criteria for class membership \cite{M21P}, as opposed to situations involving well-defined random variables \cite{B14L}. The versatility of possibility theory consequently extends its applicability across a broad spectrum of tasks, ranging from pattern recognition and data science to various other domains \cite{J24D,U24F}.
\newline
\newline
\noindent
Possibility theory, introduced by Zadeh within the framework of fuzzy set theory \cite{Z65F}, allows inference in variational inference (VI) to be reformulated using possibility-based mechanisms \cite{C95S}. Like probability theory, it establishes a formal ordering of preferences, enabling belief representation, prioritization of plausible hypotheses, resolution of inconsistencies, and identification of salient relationships \cite{V06T}. To unify probability and possibility, the possibility–probability consistency principle \cite{Z99F} provides a basis for calculating and updating possibility distributions, characterizing the coherence or absence of contradiction between information derived from both frameworks \cite{D87O}.
\newline
\newline
\noindent
Building on this foundation, \cite{H19O}
introduces a theoretical framework for constructing a joint possibility distribution, paving the way for defining a multivariate possibility distribution in terms of the dimensions of the space where uncertainty is investigated.
\newline
\newline
\noindent
To the best of our knowledge, the concept of possibility, described as an imprecise probability distribution \cite{W91S} and representing a more common occurrence in reality, has received limited attention in this domain.
While there have been some possibility-based contributions \cite{C95S,C20M,D06I,H22R,T24A} that address the problem by evaluating the plausible values of parameters, our method takes a different approach by using a simultaneous \textbf{"one vs. rest" evaluating technique} and building a \textbf{contrastive learning} function \cite{H06D} to achieve the most flexible distribution.
\newline
\newline
\noindent
The proposed framework extends variational inference (VI) by incorporating the concept of possibility. To this end, the \textbf{posterior} function in the original VI is replaced by another function, namely the \textbf{likelihood ratio} function \cite{N28O}. The advantages of this approach are explained below.

\begin{itemize}
    \item Firstly, the likelihood ratio, explicitly revised, considers both the positive and negative facets of decision-making, fostering greater confidence in the process and serving as a \textbf{leading indicator}. Its significance becomes apparent in a framework grounded in possibility, where the distinction between higher possibility and greater likelihood becomes crucial, as higher possibility does not necessarily imply a greater likelihood of occurrence.
    \item Secondly, deploying the likelihood ratio results in a contrastive learning function as an optimization objective, providing a \textbf{quantitative metric}. Compared to the original VI model, this function not only quantifies uncertainty with greater precision but also selects a function that more closely aligns with the true distribution within a narrower interval, thereby contributing to more reliable and robust model outcomes.
    \item Lastly, when deploying the Principle of Insufficient Reason \cite{B96W}, also known as the Principle of Indifference \cite{K21T}, our approach allows the use of the likelihood distribution in the likelihood ratio function instead of the posterior, wherever deemed necessary. It is noteworthy that, unlike the posterior, the choice of the likelihood distribution is within our control, and it extends the \textbf{applicability} of the model, specifically in the face of imbalanced data.
\end{itemize}

\noindent
The framework clearly holds all the properties of 
a Robust Design Method (RDM) \cite{E13A}. These aspects, however, will be examined and discussed in greater detail in sections (\ref{Exploring Likelihood Ratio as an Alternative}), (\ref{Contrastive Learning as a Model Objective Function}), and (\ref{The Final Edge in Practical Application}). 
\newline
\newline
\noindent
This paper is structured as follows: We begin with a comprehensive model description that explores the relevant background essential to our study. Next, we examine the integration of the likelihood ratio within the framework of BI, emphasizing its role in enhancing analytical rigor. Subsequently, we present a novel possibility-based framework aimed at further refining our methodology. Finally, we conclude by summarizing the proposed framework and outlining potential directions for future research.

\section{Model Description}
\subsection{Probability-based Inference}
\subsubsection{Background: Inference Challenges and Navigating Towards Variational Optimization}\label{background}
\leavevmode\\
\newline
\noindent
Consider a fully Bayesian model where each parameter is assigned a \textbf{prior} distribution, and the model incorporates both parameters and latent variables. The collection of all latent variables and parameters is represented as $\mathrm{Z} = \{\boldsymbol{z}_{1}, \hdots,\boldsymbol{z}_{M}\}$. Concurrently, the set of i.i.d observed varaibles(data) is denoted as $\mathrm{X} = \{\boldsymbol{x}_{1}, \hdots,\boldsymbol{x}_{n}\}$.
\newline
\newline
\noindent
\textbf{Note: Throughout this contribution, all vectors are represented in bold font, i.e.,} $\overrightarrow {x} := \boldsymbol{x}$\textbf{.}
\newline
\newline
\noindent
When establishing the joint distribution $p(\boldsymbol{x}, \boldsymbol{z})$, which includes both observed and latent variables, the distribution of the observed variables is derived through a process called marginalization.

\begin{equation} \label{evidence}
    p(\boldsymbol{x}) = \sum_{\boldsymbol{z} \in \mathrm{Z}} p(\boldsymbol{x}, \boldsymbol{z})
\end{equation}

\noindent
This technique allows intricate marginal distributions over observed variables to be expressed in terms 
of more tractable joint distributions across the extended space of both observed and latent variables.
It is important to note that $p(\boldsymbol{x})$ is referred to as \textbf{evidence}, since a higher marginal value suggests a better model fit.
Given our representation of $p(\boldsymbol{x})$, it logically follows that for each observation $\boldsymbol{x}_{i} = ({x}_{i1}, \hdots, {x}_{ir}) \in \mathrm{X}$, there exists a corresponding latent variable $\boldsymbol{z}_{i} = ({z}_{i1}, \hdots, {z}_{is}) \in \mathrm{Z}$. In the following, we will consistently rely on these assumptions.
\newline
\newline
\noindent
Considering that the primary objective of any probabilistic model is to assess the \textbf{posterior} distribution $p(\boldsymbol{z}_{n} \mid \boldsymbol{x}_{n})$, VI stands as no exception to this fundamental task. Based on Bayes' theorem the posterior is given by;

\begin{equation} \label{Bayes theorem}
    p(\boldsymbol{z}_{n} \mid \boldsymbol{x}_{n}) = \frac{p(\boldsymbol{x}_{n} \mid \boldsymbol{z}_{n}) \cdot p(\boldsymbol{z}_{n})}{p(\boldsymbol{x}_{n})}
\end{equation}

\noindent
with $p(\boldsymbol{x}_{n} \mid \boldsymbol{z}_{n})$ representing the \textbf{likelihood} distribution and $p(\boldsymbol{z}_{n})$ denoting the \textbf{prior}. 
The objective is to maximize the likelihood distribution.
In the equation (\ref{Bayes theorem}), the computation of $p(\boldsymbol{x}_{n})$ is either infeasible or demands exponential time to achieve a closed form. This highlights the challenge in inference approximation for such models. 
It is important to note that probabilistic inference, including approximate inference, is a computationally challenging task, classified as $\mathrm{NP-}$hard \cite{J00T}.
\newline
\newline
\noindent
To tackle the challenge of the inference problem, VI suggests utilizing a designated family of densities denoted as $\mathcal{Q}$ for the latent variables. The aim is to select the most suitable candidate from this family, representing the best estimate for the true posterior distribution.
It's noteworthy that any $q(\mathrm{Z}) \in \mathcal{Q}$ is considered a candidate approximation for the conditional probability $p(\mathrm{Z} \mid \mathrm{X})$.
The search for the optimal match, representing the closest approximation to the conditional probability, involves using the KL divergence \cite{C75I}. Consequently, the challenge of approximating inference is reframed as an optimization problem, where the complexity of the family $\mathcal{Q}$ determines the difficulty of the task.

\begin{equation} \label{the best candidate}
    q^{*}(\mathrm{Z}) = \arg\min_{q \in \mathcal{Q}} \mathrm{KL}(q(\mathrm{Z})\mid\mid p(\mathrm{Z}\mid \mathrm{X})) 
\end{equation}

\noindent
In equation (\ref{the best candidate}), $\mathrm{KL}(\bullet \mid\mid \bullet)$ denotes the KL divergence as follows. 

\begin{equation} \label{KL divergence}
    \mathrm{KL}(q(\mathrm{Z})\mid\mid p(\mathrm{Z}\mid \mathrm{X})) = - \sum_    {i=1}^{N} q(\boldsymbol{z}_{i}) \cdot \log{\frac{p(\boldsymbol{z}_{i}\mid \boldsymbol{x}_{i}))}{q(\boldsymbol{z}_{i})}} 
\end{equation}

\noindent
Upon closer examination of (\ref{KL divergence}), it becomes apparent that;

\begin{equation} \label{Kuhlback-Leibler divergence-expansion}
    \mathrm{KL}(q(\mathrm{Z})\mid\mid p(\mathrm{Z}\mid \mathrm{X})) =
    \mathrm{E} (\log(q(\mathrm{Z})) - \mathrm{E} (\log(p(\mathrm{Z}\mid \mathrm{X})))
\end{equation}

\noindent
with all expectations computed with respect to $q(\mathrm{Z})$. 
The objective function represented by equation (\ref{Kuhlback-Leibler divergence-expansion}) is not computable due to its dependence on $p(\mathrm{X})$, as elucidated in the subsequent equation. 

\begin{equation} \label{KL divergence-ultimate}
    \mathrm{KL}(q(\mathrm{Z})\mid\mid p(\mathrm{Z}\mid \mathrm{X})) =
    \mathrm{E} (\log(q(\mathrm{Z}))) - \mathrm{E} (\log(p(\mathrm{Z}, \mathrm{X}))) + \log(p(\mathrm{X})
\end{equation}

\noindent
Given the computational challenges and time-intensive nature associated with equation (\ref{KL divergence-ultimate}), the optimization process focuses on its equivalent but more manageable alternative as the objective function.

\begin{equation} \label{ELBO}
    \mathrm{ELBO}(q(\mathrm{Z})) = \mathrm{E} (\log(p(\mathrm{Z}, \mathrm{X}))) - \mathrm{E} (\log(q(\mathrm{Z})))
\end{equation}

\noindent
The present equation (\ref{ELBO}) is referred to as the Evidence Lower Bound (ELBO). Closer examination reveals that equation (\ref{ELBO}) can also be written as the following.

\begin{equation} \label{ELBO constant}
    \mathrm{ELBO}(q(\mathrm{Z})) = \log(p(\mathrm{X})) - \mathrm{KL}(q(\mathrm{Z})\mid\mid p(\mathrm{Z}\mid \mathrm{X}))
\end{equation}

\noindent
As seen in equation (\ref{ELBO constant}), $\log(p(\mathrm{X}))$ is independent of $q(\mathrm{Z})$ and therefore remains constant.
As a result, maximizing the ELBO leads to minimizing the (\ref{KL divergence}).
\newline
\newline
\noindent
In the work by \cite{k13A}, they introduce an estimator for the ELBO known as Stochastic Gradient Variational Bayes (SGVB). This method facilitates more efficient approximate inference and learning in probabilistic models. Notably, SGVB offers the advantage of improved lower bound estimation, which can be optimized using any standard stochastic gradient method.
\newline
\newline
\noindent
Given this background, we aim to further enhance the concept of VI by incorporating the notion of possibility. Initially, we confine our discussion to probability-based considerations. After establishing a framework within this context, we gradually relax these conditions and transition towards exploring the concept of possibility.  

\subsubsection{Exploring Likelihood Ratio as an Alternative to Posterior in BI} \label{Exploring Likelihood Ratio as an Alternative}
\leavevmode\\
\newline
\noindent
In section \ref{background}, the concepts of the posterior distribution (\ref{Bayes theorem}) and the evidence $p(\boldsymbol{x})$ were discussed. The posterior distribution plays a crucial role in acquiring model parameters based on observed data, facilitating model selection. On the other hand, the evidence proves invaluable within the domain of model comparison, providing a means to evaluate the efficacy of each model in explaining the observed data. Consequently, this prompts a reformulation of the inference problem as an optimization task by leveraging KL Divergence.
\newline
\newline
\noindent
Building on the principles of Robust Soft Learning Vector Quantization (RSLVQ) \cite{S03S}, a probabilistic variant of Learning Vector Quantization (LVQ) \cite{K95L}, we introduce the following revised \textbf{likelihood ratio} function, denoted by \textbf{LR}:

\begin{equation} \label{likelihood ratio}
    \mathrm{LR}(\boldsymbol{x}_{n}, \boldsymbol{z}_{n}) = \frac{p(\boldsymbol{z}_{n} \mid \boldsymbol{x}_{n})}{p(\overline{\boldsymbol{z}}_{n} \mid \boldsymbol{x}_{n})}
\end{equation}

\noindent
where 
$p(\boldsymbol{z}_{n} \mid \boldsymbol{x}_{n})$
represents a measure of certainty regarding the latent variable given the observed data. To clarify, when the conditional probability assigns a bigger probability mass to a specific value of the latent variable, it indicates higher confidence or belief in that particular choice given the observed data.
\newline
\newline
\noindent
In contrast, $p(\overline{\boldsymbol{z}}_{n} \mid \boldsymbol{x}_{n}) = \sum_{i \neq n} p(\boldsymbol{z}_{i} \mid \boldsymbol{x}_{n})$, as the cumulative probability associated with latent variable choices excluding $\boldsymbol{z}_{n}$, establishes a measure of uncertainty due to its inverse relationship with the likelihood ratio as defined in (\ref{likelihood ratio}). 
It's important to emphasize that the likelihood ratio (\ref{likelihood ratio}) is a monotonically increasing function, reaching its maximum at $p(\boldsymbol{z}^{*} \mid \boldsymbol{x}) = \max_{i} \{p(\boldsymbol{z}_{i} \mid \boldsymbol{x})\}$.
\newline
\newline
\noindent
A comparison between the posterior (\ref{Bayes theorem}) and the likelihood ratio (\ref{likelihood ratio}) reveals that \textbf{our choice of decision function leads to the immediate omission of the evidence} (\ref{evidence}), \textbf{resulting in hassle-free calculations which is the main advantage.}
For clarification, the calculation of $p(\overline{\boldsymbol{z}}_{n} \mid \boldsymbol{x}_{n}) = \sum_{i \neq n} p(\boldsymbol{z}_{i} \mid \boldsymbol{x}_{n})$ is limited by the number of latent variables. It is important to note that the subscript of any latent variable refers to its relationship with the corresponding observed data, rather than to the variable's index.
\newline
\newline
\noindent
Following the introduction of the likelihood ratio and its crucial advantage, it is time to explore its role in the objective function and the benefit of integrating it into the existing method.   

\subsubsection{Contrastive Learning as a Model Objective Function} \label{Contrastive Learning as a Model Objective Function}
\leavevmode\\
\newline
\noindent
As in the original VI, we aim to identify the most suitable surrogate probability function $q(\bullet)$ to substitute for $p(\bullet \mid \bullet)$. However, $\mathrm{LR}(\bullet, \bullet)$ (\ref{likelihood ratio})fails to meet the criteria of a probability function.
This limitation prevents us from using divergence directly to quantify the disparity between $q(\bullet)$ and $p(\bullet \mid \bullet)$. Therefore, we must define the subsequent \textbf{local loss function}, which is denoted as \textbf{loss}, to make further progress.

\begin{equation} \label{loss}
    \mathrm{loss}(\boldsymbol{x}_{n}, \boldsymbol{z}_{n}) = - q(\boldsymbol{z}_{n}) \cdot \log \Big(\mathrm{LR}( \boldsymbol{x}_{n}, \boldsymbol{z}_{n}) \Big) 
\end{equation}

\noindent
To advance, we substitute the expanded version of the likelihood ratio (\ref{likelihood ratio}) into (\ref{loss}).

\begin{align*}
    \mathrm{loss}(\boldsymbol{x}_{n}, \boldsymbol{z}_{n}) = & -q(\boldsymbol{z}_{n}) \cdot \log \Big(\frac{p(\boldsymbol{z}_{n} \mid \boldsymbol{x}_{n})}{p(\overline{\boldsymbol{z}}_{n} \mid \boldsymbol{x}_{n})} \Big) \\
    = & -q(\boldsymbol{z}_{n}) \cdot \log \Big(\frac{p(\boldsymbol{z}_{n} \mid \boldsymbol{x}_{n})}{p(\overline{\boldsymbol{z}}_{n} \mid \boldsymbol{x}_{n})} \cdot \frac{q(\boldsymbol{z}_{n})}{q(\boldsymbol{z}_{n})} \Big) \\
    = & -q(\boldsymbol{z}_{n}) \cdot \log \Big(\frac{p(\boldsymbol{z}_{n} \mid \boldsymbol{x}_{n})}{q(\boldsymbol{z}_{n})} \Big) +  q(\boldsymbol{z}_{n}) \cdot \log \Big(\frac{p(\overline{\boldsymbol{z}}_{n} \mid \boldsymbol{x}_{n})}{q(\boldsymbol{z}_{n})} \Big)\\
\end{align*}

\noindent
By summing up both sides of the aforementioned equation across all potential values of $n$, we obtain the resulting \textbf{global loss function}, referred to as \textbf{Loss}.

\begin{equation} \label{Loss}
    \mathrm{Loss}(\mathrm{X}, \mathrm{Z}) = \mathrm{KL}(q(\mathrm{Z})\mid\mid p(\mathrm{Z}\mid \mathrm{X})) - \mathrm{KL}(q(\mathrm{Z})\mid\mid p(\mathrm{\overline{Z}}\mid \mathrm{X}))
\end{equation}

\noindent
In the preceding equation (\ref{Loss}), the first and second terms on the right-hand side are interpreted as positive- and negative-reasoning, respectively.
As a result, (\ref{Loss}) is a contrastive learning function, substantiating the rationale behind the initial selection of (\ref{loss}). 
Consequently, we can formulate our objective function as follows.

\begin{equation} \label{our best candidate}
    q^{*}(\mathrm{Z}) = \arg\min_{q \in \mathcal{Q}} \mathrm{Loss}(\mathrm{X}, \mathrm{Z})
\end{equation}

\noindent
Unlike the objective function (\ref{the best candidate}) in the original VI, which considers only positive reasoning, $\mathrm{KL}(q(\mathrm{Z})\mid\mid p(\mathrm{Z}\mid \mathrm{X}))$, \textbf{the present objective function (\ref{our best candidate}) not only avoids additional computational complexities, due to the absence of the evidence, but also yields more accurate outcomes by taking into account both positive and negative reasoning.}
\newline
\newline
\noindent
In addition to the above advantages, which result in more accurate outcomes and better performance, we should bear in mind that the model is constructed through a transparent process, equipping it with a more tractable representation. As an inference model, it also provides insight into the relationships it has learned. In other words, representation, inference, and learning collectively contribute to \textbf{model-based} interpretability, as highlighted in \cite{K09P} and \cite{M19D}.
\newline
\newline
\noindent
The next section will discuss the final advantage of this approach, presenting it as the silver lining.

\subsubsection{The Final Edge in Practical Application} \label{The Final Edge in Practical Application}
\leavevmode\\
\newline
\noindent
We begin this section with the Principle of Indifference (PI) \cite{K21T}. This principle states that, in the absence of any known reason to prefer one option over its alternatives, all options should be considered equally likely based on the available knowledge. Consequently, in such a special case, applying this principle leads us to assign an equal prior probability to each class of data, i.e., $\frac{1}{n}$, where $n$ is the number of classes.
\newline
\newline
\noindent
Understanding the PI and investigating (\ref{Loss}) more closely reveals an additional advantage of deploying such an objective function.
To elaborate:

\begin{align*}
    &\mathrm{KL}(q(\mathrm{Z})\mid\mid p(\mathrm{Z}\mid \mathrm{X})) - \mathrm{KL}(q(\mathrm{Z})\mid\mid p(\mathrm{\overline{Z}}\mid \mathrm{X})) =\\
    &\sum_{n}  -q(\boldsymbol{z}_{n}) \cdot \log \frac{p(\boldsymbol{z}_{n} \mid \boldsymbol{x}_{n})}{p(\overline{\boldsymbol{z}}_{n} \mid \boldsymbol{x}_{n})} = \\
    &\sum_{n}  -q(\boldsymbol{z}_{n}) \cdot \Big( \log (p(\boldsymbol{z}_{n} \mid \boldsymbol{x}_{n}) - \log (p(\overline{\boldsymbol{z}}_{n} \mid \boldsymbol{x}_{n}) \Big)= \\
    &\sum_{n}  -q(\boldsymbol{z}_{n}) \cdot \Big( \log (p(\boldsymbol{x}_{n}, \boldsymbol{z}_{n}) - \log p(\boldsymbol{x}_{n}) - \log (p(\boldsymbol{x}_{n}, \overline{\boldsymbol{z}}_{n}) + \log p(\boldsymbol{x}_{n}) \Big)= \\
    &\sum_{n}  -q(\boldsymbol{z}_{n}) \cdot \log \frac{p(\boldsymbol{x}_{n}, \boldsymbol{z}_{n})}{p(\boldsymbol{x}_{n}, \overline{\boldsymbol{z}}_{n})} \\
    &\sum_{n}  -q(\boldsymbol{z}_{n}) \cdot \log \Big(\frac{p(\boldsymbol{x}_{n}, \boldsymbol{z}_{n})}{p(\boldsymbol{x}_{n}, \overline{\boldsymbol{z}}_{n})} \cdot \frac{\frac{1}{n}}{\frac{1}{n}} \Big)= \\
    &\sum_{n}  -q(\boldsymbol{z}_{n}) \cdot \log \frac{p(\boldsymbol{x}_{n} \mid \boldsymbol{z}_{n})}{p(\boldsymbol{x}_{n} \mid \overline{\boldsymbol{z}}_{n})}
\end{align*}

\noindent
in other words, the \textbf{posterior} density in the initial objective function (\ref{likelihood ratio}) can be replaced with the \textbf{likelihood} density, leading to the reformulation of the subsequent Loss.

\begin{equation} \label{PI-based Loss}
    \mathrm{Loss}(\mathrm{X}, \mathrm{Z}) = \sum_{n}  -q(\boldsymbol{z}_{n}) \cdot \log \frac{p(\boldsymbol{x}_{n} \mid \boldsymbol{z}_{n})}{p(\boldsymbol{x}_{n} \mid \overline{\boldsymbol{z}}_{n})}    
\end{equation}

\noindent
Although the PI is dismissed by some as a standard of ideal rationality \cite{E19P}, the current loss function (\ref{PI-based Loss}) possesses its own independent and unique merits, particularly in addressing the challenges of learning from imbalanced data. To elaborate, standard methods not only introduce bias in favor of the majority class but also struggle to distinguish between minority and noisy data \cite{D18E}.
\newline
\newline
\noindent
Along with deploying effective techniques such as resampling \cite{T13H} and generating synthetic samples \cite{H08A}, \cite{L17D}, \cite{W17V} to address real-world challenges, the utilization of (\ref{PI-based Loss}) complements efforts to advance ML methods,  highlighting its crucial role in the practical application of imbalanced learning.
\newline
\newline
\noindent
The discussion so far has focused on the choice of the likelihood ratio and the corresponding contrastive-learning objective function within the probability-based framework, highlighting its clear advantages over the original VI.
Nevertheless, while probability has long been central to decision-making, some contend that much of the information underpinning these decisions is inherently possibilistic \cite{Z99F}.
As a result, exploring a possibility-based framework becomes crucial to broadening the applicability of our method across diverse scenarios and achieving greater accuracy and reliability in outcomes.
\newline
\newline
\noindent
The subsequent section is entirely dedicated to relaxing some constraints of probability and shifting towards a possibility-based framework.

\subsection{Possibility Theory: A Complementary Approach to Probability-based Inference} 
In statistical inference, population parameters are estimated from sample statistics, as observing the entire population is usually impractical. However, within this inductive framework, the exact values of the parameters cannot be determined with complete certainty due to the inherent limitations of sample-based inference.
\newline
\newline
\noindent
BI addresses this issue by providing an order of preference among plausible parameter values, based on both prior information and the likelihood of the observed data. These preferences are represented using probability functions, which allow us to quantify uncertainty.
\newline
\newline
\noindent
As has already been argued, probability measures may not always be the best tool for inductive reasoning \cite{P87W}. In such cases, the possibility measure, an alternative framework that relaxes some of the limitations of probability and has proven to be a viable alternative in statistical inference \cite{C95S}, can be deployed.

\subsubsection{Comparing Disparity Based on Generalized Divergences}
\leavevmode\\
\newline
\noindent
In the context of probability theory, we encounter a concept known as a $\sigma$-algebra (or $\sigma$-field) \cite{B14L}, which essentially encompasses the set of all conceivable outcomes $\mathrm{X}$ associated with any given event. The fundamental significance of $\sigma$-algebra lies in their pivotal role in establishing a measure for each plausible outcome, denoted as $\boldsymbol{x}_{i} \in \mathrm{X}$, through a density function $p(\boldsymbol{x}_{i})$. Importantly, $\sum_{i} p(\boldsymbol{x}_{i}) = 1$. 
However, in practical situations involving incomplete information, the collection of conceivable outcomes does not satisfy the criteria for a $\sigma$-algebra. To clarify, unlike probabilities, the degree of membership [71], which represents a possibility distribution within the interval $[0, 1]$, is not subject to the same normalization constraint \cite{G76T}. In other words, $\sum_{i} p(\boldsymbol{x}_{i}) \neq 1$.
\newline
\newline
\noindent
With particular emphasis on the following assumptions, 

\begin{itemize} 
    \item $\forall{i}; \hspace{1mm} p(\boldsymbol{x}_{n}, \boldsymbol{z}_{i}) \geq 0$ represents a non-negative possibility measure.     \item  $\sum_{n} p(\boldsymbol{x}_{n}) \neq 1$.
   
\end{itemize}

\noindent
we leverage the equation, $\mathrm{Loss}(\mathrm{X}, \mathrm{Z}) = \sum_{n}  -q(\boldsymbol{z}_{n}) \cdot \log \frac{p(\boldsymbol{x}_{n}, \boldsymbol{z}_{n})}{p(\boldsymbol{x}_{n}, \overline{\boldsymbol{z}}_{n})}$.
, (\ref{The Final Edge in Practical Application}), as the foundation of the possibility-based scheme. 
It is important to note that $q(\bullet)$ is also a possibility distribution. Consequently, the KL divergence in equation (\ref{Loss}) is unable to effectively measure the dissimilarity between the non-normalized 
$q(\bullet)$ and $p(\bullet, \bullet)$. Therefore, we deploy extended $\alpha$-divergence \cite{Z97M}, a parameterized form of the Csiszár-Morimoto and Bregman divergences, as detailed below.

\begin{equation} \label{alpha-divergence}
    \mathrm{D}^{\alpha}(q\mid\mid p) = -\frac{1}{\alpha \cdot (\alpha-1)} \cdot \sum_{n} q_{n}^{\alpha} \cdot p_{n}^{1-\alpha} - \alpha \cdot q_{n} + (\alpha-1) \cdot p_{n} \hspace{1mm}; \hspace{1mm} \alpha \in \mathbb{R}-\{0,1\}
\end{equation}

\noindent
It is important to note that the aforementioned divergence enhances both the flexibility and efficiency of algorithms such as expectation-maximization, thereby improving their robustness \cite{C10F}.
Furthermore, a closer investigation shows as $\alpha \xrightarrow{} 1$, the $\alpha$-divergence converges to the \textbf{generalized} KL divergence \cite{Z95B}, $\mathrm{KL}_{g}(\bullet \mid\mid \bullet)$.

\begin{equation} \label{generalized KL}
    \lim_{\alpha \xrightarrow{} 1} \mathrm{D}^{\alpha}(q\mid\mid p) = - \sum_{n} q_{n} \cdot \log (\frac{p_{n}}{q_{n}}) - q_{n} + p_{n} = \mathrm{KL}_{g}(q\mid\mid p)
\end{equation}
\noindent
When (\ref{generalized KL}) is restricted to normalized positive measures, we obtain the \textbf{standard} KL divergence.

\begin{equation}
    \lim_{\alpha \xrightarrow{} 1} \mathrm{D}^{\alpha}(q\mid\mid p) = - \sum_{n} q_{n} \cdot \log (\frac{p_{n}}{q_{n}}) = \mathrm{KL}(q\mid\mid p)
\end{equation}

\noindent
By deploying the $\alpha$-divergence, the Loss (\ref{Loss}) can be expressed as follows.

\begin{equation} \label{possibility-based loss function}
    \mathrm{Loss}_{g}(\mathrm{X}, \mathrm{Z}) = \mathrm{D}^{\alpha}(q(\mathrm{Z})\mid\mid p(\mathrm{Z}\mid \mathrm{X})) - \mathrm{D}^{\alpha}(q(\mathrm{Z})\mid\mid p(\mathrm{\overline{Z}}\mid \mathrm{X}))
\end{equation}

\noindent
Finally the substitution of the \textbf{generalized Loss}, denoted by ($\mathrm{\textbf{Loss}}_{\textbf{g}}$) (\ref{possibility-based loss function}) into the objective function (\ref{our best candidate}) yields a revised objective function, thereby establishing a robust possibility-based inference framework. 

\section{Discussion}
Undoubtedly, significant empirical work is needed to deepen our understanding of this mathematical model. Although the application of the likelihood ratio in the context of BI is novel, its deployment in other studies \cite{M22P,X20T} demonstrates the feasibility of the approach and suggests that concerns about the computational aspects and efficiency of the inference process in this model may be unwarranted. Furthermore, these studies provide evidence of the broad applicability of the likelihood ratio.

\section{Conclusion}
\noindent
In this paper, we introduce a comprehensive mathematical model for a novel inductive inference framework. Recognizing the crucial role of VI and the computational challenges inherent in Bayesian ML models, we propose a methodical approach that integrates a possibility-based decision function into VI. This results in a robust and reliable platform that enhances decision-making by providing more interpretable support.
\newline
\newline
\noindent
On one hand, restricting the model to probability-based measures makes it suitable for predictive inference; on the other hand, incorporating possibility measures facilitates estimation inference \cite{H16L}. As a consequence, this framework, centered around a contrastive learning function, not only models uncertainty but also enables informed decision-making across various domains, including computer vision, natural language processing, and reinforcement learning.
\newline
\newline
\noindent
Specifically, the flexibility of the contrastive learning function, which utilizes likelihood density instead of the standard posterior, contributes to deeper insights into the nature of imbalanced learning at both the data and algorithmic levels.
\newline
\newline
\noindent
Although the application of the likelihood ratio in the context of BI is novel, its application in other studies \cite{M22P,X20T} provides evidence of its practical feasibility in related settings. However, significant empirical work remains necessary to deepen our understanding of the proposed mathematical model, particularly with respect to the computational aspects and efficiency of the inference process.
\newline
\newline
\noindent
In conclusion, this framework opens up new avenues for future research, including empirical investigations of its computational performance, robustness, and applicability across diverse domains and datasets.

\section*{Acknowledgements}
This work was conducted at the Federal Institute for Occupational Safety and Health (BAuA Dresden) as part of the junior research group “Artificial Intelligence (AI) in a Safe and Healthy Working Environment”, with support from the German Federal Ministry of Labour and Social Affairs. The authors gratefully acknowledge the institute's permission to make this work publicly available.

\bibliographystyle{plain}
\bibliography{References}

\end{document}